\documentclass[conference]{IEEEtran}
\IEEEoverridecommandlockouts
\usepackage{cite}
\usepackage{amsmath,amssymb,amsfonts}
\usepackage{algorithmicx}
\usepackage{graphicx}
\usepackage{textcomp}
\usepackage{comment}
\usepackage{subcaption}
\usepackage{booktabs}
\usepackage{xcolor}
\usepackage{algorithm, algpseudocode}
\usepackage{physics}
\def\BibTeX{{\rm B\kern-.05em{\sc i\kern-.025em b}\kern-.08em
    T\kern-.1667em\lower.7ex\hbox{E}\kern-.125emX}}
\begin{document}

\author{\IEEEauthorblockN{Arwin Gansekoele\IEEEauthorrefmark{1}\IEEEauthorrefmark{2}, Tycho Bot\IEEEauthorrefmark{3}, Rob van der Mei\IEEEauthorrefmark{1}\IEEEauthorrefmark{2}, Sandjai Bhulai\IEEEauthorrefmark{2} and Mark Hoogendoorn\IEEEauthorrefmark{2}}
\IEEEauthorblockA{\IEEEauthorrefmark{1}Centrum Wiskunde \& Informatica, Amsterdam, the Netherlands}
\IEEEauthorblockA{\IEEEauthorrefmark{2}Vrije Universiteit, Amsterdam, the Netherlands}
\IEEEauthorblockA{\IEEEauthorrefmark{3}Ministry of Defence, The Hague, the Netherlands}
\IEEEauthorblockA{Email: arwin.gansekoele@cwi.nl}
}

\title{Unveiling the Potential: Harnessing Deep Metric Learning to Circumvent Video Streaming Encryption}


\maketitle

\begin{abstract}
    Encryption on the internet with the shift to HTTPS has been an important step to improve the privacy of internet users. However, there is an increasing body of work about extracting information from encrypted internet traffic without having to decrypt it. Such attacks bypass security guarantees assumed to be given by HTTPS and thus need to be understood. Prior works showed that the variable bitrates of video streams are sufficient to identify which video someone is watching. These works generally have to make trade-offs in aspects such as accuracy, scalability, robustness, etc. These trade-offs complicate the practical use of these attacks. To that end, we propose a deep metric learning framework based on the \emph{triplet loss} method.

Through this framework, we achieve robust, generalizable, scalable and transferable encrypted video stream detection. First, the triplet loss is better able to deal with video streams not seen during training. Second, our approach can accurately classify videos not seen during training. Third, we show that our method scales well to a dataset of over $1000$ videos. Finally, we show that a model trained on video streams over Chrome can also classify streams over Firefox. Our results suggest that this side-channel attack is more broadly applicable than originally thought. We provide our code alongside a diverse and up-to-date dataset for future research.
\end{abstract}


\section{Introduction}

The Internet has become an indispensable part of everyday life in modern society. Being able to interact with family and friends in the blink of an eye, irrespective of their location is among the greatest accomplishments of the past century. A vital component in the pipeline to sharing information over the internet is the \emph{Hypertext Transfer Protocol} (HTTP).

HTTP was not designed initially with any security features, however. If a third party wanted to see the content of someone’s actions on the internet, it would be reasonably simple to do so. Such a third party could even pretend to be one of the communicating parties, a \emph{man-in-the-middle} (MITM) attack. With that risk in mind, a secure version of HTTP called \emph{HyperText Transfer Protocol Secure} (HTTPS) was proposed. This version added encryption and verification to the existing HTTP protocol.

This transition had major implications for automated packet inspection methods. Without encryption, connections could be inspected automatically using connection parameters to identify the traffic, e.g., video, audio, chat etc.\cite{dainotti2012in, valenti2013dtma, salman2020at}. This transition is thus great news for the end user as encryption gives a strong guarantee of privacy. However, it also complicates the work of internet service providers (ISPs), as well as giving cyber criminals more leeway to go undetected. Clearly, there are parties that would benefit from bypassing HTTPS. It is thus important to understand the weaknesses in the HTTPS protocol to be able to patch them. To this end, a new form of side-channel attacks has been researched recently \cite{velan2015ijnm}. While it is nearly impossible to decrypt the content of an interaction, the interaction itself still contains information. 

One instance of such connections is video streaming. The majority of video streams are transmitted using either the \emph{HTTP live streaming} (HLS) or \emph{dynamic adaptive streaming over HTTP} (MPEG-DASH) protocol. These enable HTTP(S) servers to deliver video content adaptively based on users’ preferences and internet availability. As such, they can identify the network conditions of users on the fly and send segments of appropriate quality. However, multiple works \cite{dubin2017itifs, schuster, reed201621iaccncc, reed2017psacdasp, gu2018, dias2019cn, wu2020ii2-icccwiwa, bae2022 ,wu2023} have shown that these protocols leak information such that a third party could identify which video is being watched when matched against a set of known videos. These range from approaches based on network statistics \cite{reed201621iaccncc, reed2017psacdasp, wu2020ii2-icccwiwa}, manual feature extraction with traditional machine learning (ML) methods\cite{dias2019cn, dubin2017itifs} and deep neural networks (DNNs)\cite{schuster}.

We find the latter two of these approaches to have their own strengths and weaknesses. A traditional ML approach such as the \emph{k-nearest neighbor} (kNN) algorithm is cheap to extend with new classes but is not as accurate as DNNs. In turn, DNNs are expensive to retrain. Furthermore, no work has yet proposed a method to deal with \emph{out-of-distribution} (OOD) data. This is data not seen during training but encountered at test time. Finally, no learning-based method is currently capable of transferring across different settings. Browsers can have different implementations of the MPEG-DASH controller, for example, which means that the same video watched on different browsers can result in entirely different streams. To address these issues, we propose a triplet loss approach with an extension we call \emph{outlier leveraging} (OL). Our approach combines the strengths of both classes of methods while also being better able to deal with OOD data. More specifically, we make the following contributions.

\begin{enumerate}

    \item \emph{Robust}. We show that triplet loss models are significantly more robust in the presence of OOD video streams than current state-of-the-art models. The addition of OL further improves robustness when OOD streams are dissimilar from the streams trained on.

    \item \emph{Generalizable}. We find that the triplet loss model allows the inclusion of new classes without retraining the model. We empirically show that our method achieves this with high accuracy.

    \item \emph{Scalable}. We demonstrate that our method scales well when the number of videos to detect is large and the number of streams available is small.

    \item \emph{Transferable}. We show that our method transfers well across settings. A model trained on video streams from Chrome is also useful for classifying video streams from Firefox.
\end{enumerate}

In practice, our work suggests the possibility of bypassing encryption on a large scale for video streaming. 

\section{Related Work}

Through the challenges defined above, we identify three related areas of research. The main area of research is encrypted streaming content detection, the identification of which videos correspond to which encrypted video streams. 

\subsection{Encrypted Video Stream Detection}

Many works have already shown that it is possible to identify the nature of encrypted network traffic \cite{dainotti2012in, valenti2013dtma, salman2020at}. While it was known that it is possible to determine the nature of traffic, Dubin et al. \cite{dubin2017itifs} were among the first to show that the connection is sufficiently distinct to determine which video is being watched. Gu et al. \cite{gu2018} created a novel fingerprinting algorithm to identify variable bit-rate video streams. Wu et al. \cite{wu2020ii2-icccwiwa} proposed a similar method but aligned the sequences instead of treating them as sets and applying ML. Reed and Klimkowski \cite{reed201621iaccncc} demonstrated that it is possible to construct fingerprints of Netflix shows statistically purely by looking at the segment sizes and bitrates. In follow-up work, Reed and Kranch \cite{reed2017psacdasp} showed that an update by Netflix did little to patch this vulnerability. Dias et al. \cite{dias2019cn} proposed a video streaming classifier based on the Naive Bayes method. Schuster et al. \cite{schuster} were among the first to use DNNs for this problem, specifically \emph{convolution neural networks} (CNNs).  Bae et al. \cite{bae2022} demonstrated the effectiveness of this attack in LTE networks. Wu et al. \cite{wu2023} showed that it is also possible to identify the resolution of a video stream. We build on this body of work through our deep metric learning approach. 


\subsection{Out-of-Distribution Detection and Open Set Recognition}

Our work also touches on the field of out-of-distribution detection (or open-set recognition). This ML field addresses the issues of models face in the presence of data they were not initially trained on. Hendrycks and Gimpel \cite{hendrycks_baseline_2016} introduced a baseline model with an evaluation suite for out-of-distribution detection. Lee et al. \cite{lee_training_2018} opted to use additional synthetic out-of-distribution samples as training input. This method worked well for synthetic out-of-distribution samples, but natural images proved problematic. Hendrycks et al. \cite{hendrycks_deep_2019} proposed the outlier exposure method, where they fine-tuned a classifier using a large dataset and showed it improves OOD detection. These works have not been applied to encrypted video streams and form the basis for the new outlier leveraging method we propose.

\subsection{Deep Metric Learning}


A weakness of the kNN approach is that applying a distance metric to compare features is not necessarily informative. Deep metric learning is an approach where neural networks are used to learn a feature transformation such that the transformed features have an informative distance metric. The most common deep metric learning approach is the triplet loss\cite{schroff20152iccvprc}. While originally thought not to be as effective as classification losses, Hermans et al. \cite{hermans2017ac} set guidelines on how to train triplet loss models and showed that they can be more effective than classification losses. Sirinam et al. \cite{sirinam2019c} approached the problem of web fingerprinting using a triplet loss-based network. They achieved promising results in adapting to new websites. Wang et al. \cite{wang2021} expanded on this by including adversarial domain adaptation to better transfer knowledge from a source dataset to a target dataset. We use these works to devise our own triplet loss-based method.

\section{Methodology}

We now explain our method to address the challenges described earlier. Recall that, in our case, we wish to determine which video was watched based on the stream we observed. A stream in this case is a time series of how many bits were received at every point in time. Every time a video is streamed, the bits received and the times at which they are received differ based on factors such as the browser or network conditions. As the stream is encrypted, we assume it is not possible to determine the video based on the content. In fact, it is important to note that the patterns captured in this side-channel attack are not based on the video content but on its MPEG-DASH segmentation. The same video hosted on different platforms would result in dissimilar streams if different parameters are used. The variance between different streams of the same video also makes it difficult to establish exact matches. That is why we use multiple different streams as samples and their corresponding videos as class labels to create a classification model. Our approach uses the triplet loss, which fulfils the same purpose but operates differently.

\subsection{Triplet Loss}

 The triplet loss is a metric learning loss approach used to teach a model to predict the similarity between samples. A neural network learns to map samples onto a metric space where samples from the same class are closer together than samples from different classes. In our case, it means that it positions streams from the same video closer together. The triplet loss is computed over a set of triplets $\mathbb{T}$ where each triplet is a combination of three unique streams: an anchor $\vb*{a}$, a positive $\vb*{p}$ and a negative $\vb*{n}$. The anchor is a stream that belongs to the same video as the positive stream, whereas the negative stream is a stream of a different video. Intuitively, the negative stream should be further away from the anchor stream than the positive stream. A margin parameter $m$ is normally included to define how much further away a negative stream should be compared to the positive stream. The goal of the triplet loss is thus to teach a neural network $f_\theta$ parameterized by parameters $\theta$ to transform some stream $x^{(i)}$ into an embedding $f_\theta(x^{(i)})$ such that any embedding $f_\theta(x^{(j)})$ of the same video lies closer to this embedding than the embeddings of streams of different videos. The 'closeness' can be defined through any differentiable distance metric. We have opted for the Euclidean distance as it is often the most effective \cite{hermans2017ac}. We define the formula for the triplet loss in Equation \ref{eq:triplet-loss}. 

This \emph{metric learning approach} differs fundamentally from classification approaches \cite{schuster}. Classification approaches teach a model of the similarity of a stream to a pre-defined video. They are directly optimised to determine which video stream belongs to which videos. Classification approaches are thus often more accurate, as they are optimised for that purpose. A metric learning approach teaches a model which streams are similar but not explicitly which videos they belong to. A major advantage of such an approach is that the resulting model is not dependent on which videos it sees during training. It can classify any arbitrary video, which has made this approach popular in other fields for n-shot learning.

\begin{equation}
\begin{aligned}
    \mathcal{L}_{triplet}(f_\theta ; \mathbb{T}; m) = \frac{1}{|\mathbb{T}|} \sum_{(a,p,n)\in \mathbb{T}} \max  \big\{0, m + \\||f_\theta(\vb*{x}^{(a)}) - f_\theta(\vb*{x}^{(p)})||_2 - ||f_\theta(\vb*{x}^{(a)}) - f_\theta(\vb*{x}^{(n)})||_2\big\}.
\end{aligned}
\label{eq:triplet-loss}
\end{equation}







We perform our triplet selection in an online fashion similar to \cite{hermans2017ac}. This means that we first generate the embeddings from a randomly sampled balanced batch. A balanced batch entails having a subset of all classes and sampling an equal amount of samples for each of those classes. After generating the embeddings, we perform triplet selection and compute the loss within this batch.

To select the triplets, we use two strategies. In the first ten epochs, we use a semi-hard negative strategy. In this strategy, we randomly select for each positive-anchor pair a random negative such that the triplet is semi-hard. A semi-hard triplet is a triplet where the distance of the anchor to the negative is larger than to the positive, but the difference is not larger than the margin parameter yet. This strategy is easy to optimize and helps avoid model collapse. After the tenth epoch, we switch to the hardest negative strategy. In this strategy, a set of triplets is formed by selecting the negative closest to the anchor for each positive-anchor pair in a batch. We found this strategy the most effective during tuning.

\subsection{Inference}

While the triplet loss defines a training procedure, we still need to define how to perform classification with the trained model. There are multiple approaches possible here, but we have opted for an approach similar to \cite{hermans2017ac}. We take the mean embedding for every class, dubbed the centroid, and treat the distance of the centroid to new data points as the class-conditional score. We define the set of embeddings $\mathbb{Z}_k$ belonging to class $k$ given some (training) dataset $\mathbb{X}$ in Equation \ref{eq:train-embeddings}.

\begin{equation}
    \mathbb{Z}_k = \left\{f_\theta(\vb*{x})\mid \left(\vb*{x}, y\right) \in \mathbb{X} , y = k \right\}.
    \label{eq:train-embeddings}
\end{equation}

We can then use these embeddings to compute the centroid $\vb*{c}_k$ for class $k$ using Equation \ref{eq:cluster-mean} below.

\begin{equation}
    \vb*{c}_{k} = \frac{1}{|\mathbb{Z}_k|}\sum_{f_\theta(\vb*{x}) \in \mathbb{Z}_k} f_\theta(\vb*{x}).
    \label{eq:cluster-mean}
\end{equation}

Using centroids is an intuitive solution, as the centroid is the point that minimizes the distance to every point in $\mathbb{Z}_k$. Given the definition in Equation \ref{eq:cluster-mean}, we define the video-conditional score of some stream $\vb*{x}$ given model $f_\theta$ with respect to video $k$ in Equation \ref{eq:class-conditional-score} below. 

\begin{equation}
    s(\vb*{x}; f_\theta, k) = \frac{\exp\left(-||\vb*{c}_k - f_\theta(\vb*{x})||_2\right)}{\sum_{j=1}^{K} \exp\left(-||\vb*{c}_j - f_\theta(\vb*{x})||_2\right)}.
    \label{eq:class-conditional-score}
\end{equation}

This corresponds to the softmax over the negative distances of $f_\theta(x)$ to all centroids. We perform prediction by taking the argmax of this score with respect to video $k$ as defined in Equation \ref{eq:prediction-label} below.

\begin{equation}
    \text{prediction}(\vb*{x}; f_\theta) = \operatorname{argmax}_{j \in \left\{1,\dots,K\right\}} s(\vb*{x}; f_\theta, j).
    \label{eq:prediction-label}
\end{equation}

We can also use the score function as a measure of confidence for the model’s prediction. A confidence measure is important when considering the OOD component of our problem. By applying a threshold to such a measure, we can determine whether a prediction is correct or not. We have defined this confidence measure in Equation \ref{eq:confidence-measure} below.

\begin{equation}
    \text{confidence}(\vb*{x}; f_\theta) = \operatorname{max}_{j\in \left\{1, \dots, K\right\}} s(\vb*{x}; f_\theta, j).
    \label{eq:confidence-measure}
\end{equation}

 \subsection{Outlier Leveraging}

Like the standard cross-entropy loss, the triplet loss is not generally trained for an open-world setting. A third party may only be interested in a handful of videos and network conditions but needs to handle them regardless. That is why we also introduce a generalisation (OL) of the triplet loss function that allows incorporating streams of OOD videos during training. The loss function $\mathcal{L}_{OL}$ is similar to Equation \ref{eq:triplet-loss} but instead of sampling negatives from other classes, they are sampled from an extra tuning dataset. Intuitively, we now also maximize the distance between the OOD streams and the anchors by using triplet loss. We combine the losses in Equation \ref{eq:combined-loss}.


\begin{equation}
    \mathcal{L} = \mathcal{L}_{triplet} + \lambda \mathcal{L}_{OL}.
    \label{eq:combined-loss}
\end{equation}

The idea behind using two separate loss functions is to keep the sampling process of video streams we wish to distinguish from each other ($\mathcal{L}_{triplet}$) separate from the sampling process of video streams we want to reject ($\mathcal{L}_{OL}$). This ensures that in every training step, the model is guaranteed to concurrently learn to be discriminative and robust. 

\section{Data Collection}

To test our method, we first gather data for our experiments. We opted to gather our own data as opposed to using datasets from prior work. We mainly do so as web technology moves quickly. Results from datasets from years ago may no longer represent the current state of affairs. For example, Google has been increasingly pushing the QUIC standard, so the dataset we gathered using Google Chrome only contains QUIC streams. An added benefit is that it gives us full control of our experimental setup and thus allowed us to construct a large, and diverse video streaming dataset. 

For our data collection, we consider one of the simplest scenarios: a third party has access to the line and can observe all incoming and outgoing traffic. This scenario applies to ISPs and government agencies. This scenario is likely generalizable to other scenarios, e.g., when considering a side-channel attack \cite{schuster} or when performing over-the-air captures \cite{li20182i1isncan}. 

To collect the data, we first gathered the video IDs we would like to observe using two approaches. The first approach consisted of scraping $20$ videos from the YouTube trending page like \cite{schuster}. The second approach entailed sampling videos pseudo-randomly from YouTube. We did so by taking a random string of size $4$ and querying $50$ videos using the YouTube API. 

For each of the collected videos, we gathered samples as follows. We opened a Selenium browser instance that opens the link to the video. We kept the browser open for 60 seconds and recorded the network traffic using tcpdump. All browser instances used AdBlock Plus to handle advertisements. We filtered out the video component by looking at the DNS requests made and then filtering on the IP address that corresponds to the video component. This is possible, as most DNS requests are not yet encrypted. In case DNS requests are not available, picking connections that transmit a lot of data over a large window of time is an effective alternative.

Given a connection, we created our features as follows. We initialized a vector of size $240$ where each element refers to a bucket with a timespan of 1/4 second in the total recording of $60$ seconds. For every 1/4 second time window, we recorded the number of bytes that were transmitted during that time. After completion, we standardized each sample individually. Standardization can be helpful when dealing with different-quality video streams. We also found it to perform better than normalization. We repeated this procedure for both the incoming and outgoing packets, resulting in the final $2\times 240$ bivariate feature vector. The gathered datasets are as follows.

\begin{enumerate}
    \item $D_{small}:$ A dataset consisting of $20$ different videos with $100$ streams per video. This dataset was collected in a manner similar to \cite{schuster}; we took $20$ videos from the trending page and streamed all of them using a Chrome browser. We use this dataset to evaluate whether our method is performant in a setting known to be 'solved.' For every video, we captured $80$ streams for training and $20$ streams for evaluation.

    \item $D_{large} \text{ and } D_{firefox}:$ Two larger datasets which comprise of $1087$ different videos with $10$ streams and $4$ streams each. We gathered them using the pseudo-random approach described earlier. We recorded them using a Chrome and Firefox browser instance respectively. We intended to capture as many different videos as we could with a small number of samples per video to evaluate the scalability of our methods. We split the data in a balanced manner with the proportions $8/2$ and $3/1$ respectively. 

    \item $D_{tune} \text{ and } D_{out}:$ Two datasets comprising $1000$ and $8000$ different videos with $1$ stream each. This dataset is comprised of different videos than in $D_{large}$ and $D_{firefox}$. The former is used for the OL tuning and the latter for the OOD evaluation. The streaming process was done using a Chrome browser.
\end{enumerate}



%



\section{Experimental Setup}

After defining the methods and data we use for our experiments, we discuss what experiments we have performed to evaluate these as well as the settings we have used to run our experiments. 

\subsection{Architecture}

We have opted for a CNN with three blocks and a fully connected hidden layer. Each block consists of two 1D convolutional filters of size $7$ followed by a ReLU activation function and batch normalization\cite{szegedy2015}. At the end of the block, max pooling is used to reduce the size of the feature map for the first two blocks with a global average pooling block used for the last block. The number of channels for the three blocks is 128, 256, and 512 respectively. A one-layer \emph{feed-forward neural network} (FFNN) with a hidden layer size of 1024 is used as the classification head. A helpful feature of a CNN is its translation invariance. This helps our network with aligning sequences, which means that our method is not limited to classifying only the start of a video.

\subsection{Hyperparameter Selection}

To optimize our network, we used the Adam \cite{kingma2017} optimizer with decoupled weight decay regularization \cite{loshchilov2019}. We used this optimizer alongside a learning rate of $3\times 10^{-4}$ and a weight decay of $0.01$. We found that the models are sufficiently regularised and that tuning does not result in drastically better models. A benefit of using default hyperparameters is that our results should be straightforward to reproduce. We note the hyperparameters for the models used to classify $D_{small}$ and $D_{large}$ in Table \ref{tab:hyperparameters-small}.

\begin{table}[hbtp]

   \centering

   \caption{The hyperparameter selection for the triplet model and benchmark (where relevant) when trained on $D_{small}$ and $D_{large}$. Where applicable, the hyperparameters are $D_{small}$/$D_{large}$ separated. Note that the videos per batch and samples per video together determine the batch size of a balanced batch. The architecture for $D_{small}$ is identical to \cite{schuster} with lower dropout values for the triplet loss model. For $D_{large}$, we use the architecture described in the previous architecture subsection.}

   \begin{tabular}{@{}ll@{}}

    \toprule

     & Selected\\

    \midrule

      Architecture &  CNN\\

      Optimiser & AdamW\\

      Number of Epochs & 250\\

      Learning Rate  &  $3\times 10^{-4}$\\

      Weight Decay & 0.01\\

      Embedding Size & $1024$\\

      Videos per Batch & 5/25\\

      Samples per Video & 25/5 \\

      Number of Runs & 10\\

      \bottomrule

    \end{tabular}

    \label{tab:hyperparameters-small}

\end{table}

\subsection{Benchmark Models} 
We evaluated our method empirically alongside two benchmark models. The first benchmark is a kNN approach applied to the input features. While similar to the approach in \cite{dubin2017itifs}, we found that their approach is not as effective as their original method when unable to filter out TLS retransmissions. We found that simply applying kNN to the input features yields surprisingly good results and thus used that approach. The kNN approach provides a lower bound for our method, as it has a trivial training time and is thus able to include new videos on the fly. We select the number of neighbors through 5-fold cross-validation. The optimal number of neighbors for the kNN approaches for all our experiments was $1$.

A second benchmark we include is the convolutional neural network approach proposed in \cite{schuster}. As this is the most accurate approach we are aware of, it is an obvious benchmark. We maintain a similar number of parameters for this benchmark compared to our method to ensure a fair comparison. For the sampling strategy, we do not have to perform triplet mining but instead sample batches of size $128$ at random. We refer to the cross-entropy model as CNN in later sections. We can compute the scores and prediction for the benchmark model by taking the max and argmax respectively of the final scores outputted by the CNN.

\subsection{Experiments}

To evaluate our experiments, we use three metrics. First, we use accuracy computed over the videos we wish to classify. This metric tells how well the model can distinguish videos of interest. Second, we compute the mAP (\emph{mean average precision}) between correctly classified videos and videos from $D_{out}$. We pose this as a binary classification problem where we wish to detect the correctly classified videos. We use the confidence score function in equation \ref{eq:confidence-measure} to determine the ranking. Third, we compute the recall at $100\%$ precision alongside the mAP. Both metrics together give an overview of how well the method performs in the presence of out-of-distribution data. We then performed four experiments to evaluate our four challenges. 

\begin{enumerate}

    \item \textbf{Robustness}. In these experiments, we evaluate how robust our models are when evaluated on $D_{small}$ and $D_{out}$. We train all models and compare them using the metrics as described in this section. We also investigate the effect of OL on performance.

    \item \textbf{Generalisability}. In these experiments, we evaluate how well our triplet-loss approach performs if we add videos that were not seen during training.

    \item \textbf{Scalability}. In these experiments, we include the dataset $D_{large}$ to evaluate whether the conclusions we found in experiments 1 and 2 apply in more challenging settings.

    \item \textbf{Transferability}. In these experiments, we evaluate whether our models trained on Chrome can be used for Firefox data as well. 

\end{enumerate}

\section{Results and Discussion}

Having defined the setup we use to evaluate our methods, we discuss the results we gathered to answer our research question. 

\subsection{Robustness}

We thus first evaluated the robustness of our methods compared to the kNN and CNN baselines described earlier. To do so, we evaluated our method on how well it classifies the streams in $D_{small}$, as well as whether it can distinguish $D_{small}$ from $D_{out}$. Otherwise, we used the setup as described in the Experimental Setup and report the result of this evaluation in Table \ref{tab:small-robustness-results}.

\begin{table}[b]
    \centering
    \caption{The results for the accuracy and robustness metrics for the models trained on $D_{small}$. The kNN model is the kNN model with a selected number of neighbors of 1. The CNN model is a softmax cross-entropy classifier as in \cite{schuster}. The triplet and triplet + OL models are our proposed triplet model as well as the outlier leveraging extension proposed. Recall is measured at $100\%$ precision.}
    \begin{tabular}{@{}llll@{}}
    \toprule
{} &       Accuracy &            mAP & Recall \\
    \midrule
    kNN &  0.99 ± 0 & 0.05 ± 0 & 0 ± 0\\
    CNN    &  1.00 ± 0.00 &  0.76 ± 0.20 &            0.19 ± 0.17 \\
    Triplet      &  1.00 ± 0.00 &  0.98 ± 0.01 &            0.59 ± 0.32 \\
    Triplet + OL &    1.00 ± 0.00 &  1.00 ± 0.00 &            0.89 ± 0.13 \\
    \bottomrule
    \end{tabular}
    \label{tab:small-robustness-results}
\end{table}

First, we found that using kNN alone is sufficient to achieve an accuracy above $99\%$. This indicates, first of all, that kNN is sufficient for a small and clean dataset of encrypted video streams to achieve a high test accuracy. However, an mAP of $5\%$ indicates that the kNN algorithm is unable to cope with new video streams; it is not robust. Based on this finding, we opted to experiment with the number of neighbors beyond the results of cross-validation. We found that increasing the number of neighbors can improve the robustness quite drastically. However, it also reduces the accuracy slightly. Consequently, to get an accurate and robust kNN algorithm, it would be necessary to tune using out-of-distribution data.

However, both the CNN and triplet loss models do not require such tuning to achieve much better open-world scores. Both the CNN and triplet loss models respectively represent big steps in terms of open-world detection performance. The CNN model already achieved an mAP of $76\%$. Intuitively, the mAP is a form of summary statistic to measure the average precision over a set of thresholds based on the recall. Using a neural network seems to benefit the robustness of the model amidst new video streams, especially as it does not require out-of-distribution data for tuning.


It was known that the CNN model is more effective than the kNN approach. Interestingly, the triplet loss model is quite a bit more robust than the CNN. The mAP of 98\% indicates that it can retrieve most of the videos we want to classify without making errors. Concretely, the recall at $100\%$ precision metric indicates that it can identify $59\%$ correctly without making any errors. The triplet loss model has access to the same data as the CNN and, unlike the kNN model, has a similar number of parameters as the CNN. The triplet loss function seems to force the backbone to extract more information from the training data. It seems to get penalized much more for being uncertain as opposed to the cross-entropy loss used by the CNN benchmark. Consequently, it is better able to recognize which video streams belong to one of the known videos and which do not. Using the triplet loss function thus results in more robust models in an open-world setting.

\subsection{Generalisability}


\begin{figure}[!h]
    \centering
    \includegraphics[width=\linewidth]{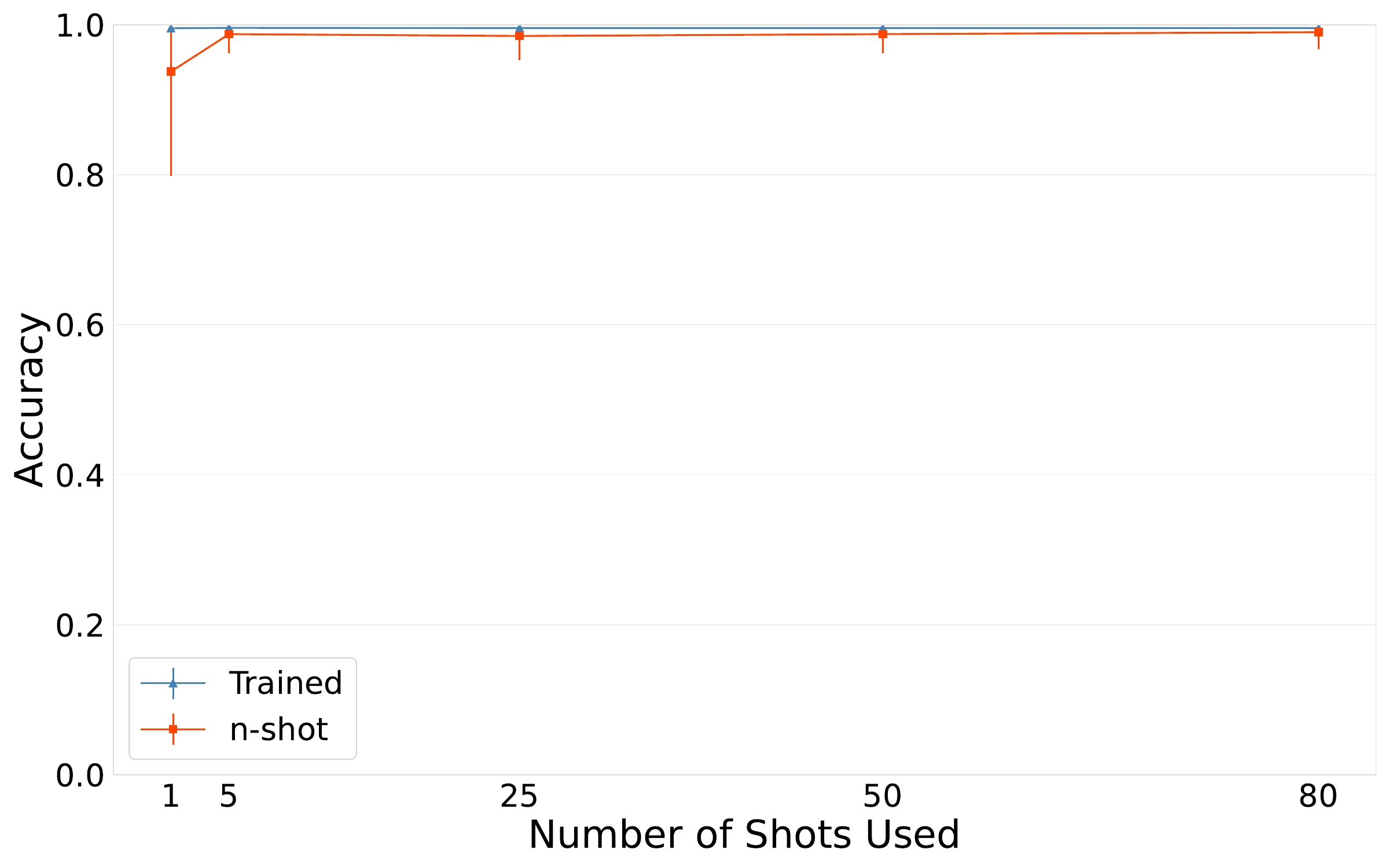}
    \caption{The accuracy for the models trained using 18 of the videos (leave out a fraction of 0.1) but where we include a variable number of shots. It seems that generating a centroid using more data has a positive effect on the accuracy.}
    \label{fig:short-nshot}
\end{figure}

We have thus shown that on the small dataset, our method is at least as accurate while being more robust than previously known methods. In theory, it also provides the same flexibility of including new videos on the fly. To test this, we randomly selected and left out $2$ of the videos in $D_{small}$, and trained on the remaining $18$ videos. We demonstrate what the impact is of incorporating $n$ shots with varying numbers for $n$ in Figure \ref{fig:short-nshot}. Interestingly, adding more shots strictly improves the performance of the model despite not using any retraining steps. When having access to the full $80$ shots, the classification of the newly added videos is almost as accurate as the triplet loss model, and more accurate than kNN. This suggests that our model may be able to not only detect OOD data but classify them as well.

\subsection{Scalability}
The dataset $D_{small}$ is fairly small and we cannot say much yet about whether our results would generalize to larger datasets. It can be argued that the dataset is also fairly simple, given how kNN is sufficient for a test accuracy of over 99\%. That is why we have opted to gather the larger dataset $D_{large}$ described earlier. This dataset consists of many different videos but few streams per video, as opposed to a few videos with many streams per video such as in $D_{small}$. This dataset gives us a better idea of what an attacker would be able to achieve at scale. We used the larger neural network architectures described in the experimental setup for these experiments. We quickly found that the small architectures were insufficient to properly learn the patterns in the data. We first evaluate the robustness again in Table \ref{tab:large-robustness-results}.

\begin{table}[t]
    \centering
    \caption{The results for the accuracy and robustness metrics for the models trained on $D_{large}$. The kNN model is the kNN model with a selected number of neighbors of 1. The CNN model is a softmax cross-entropy classifier as in \cite{schuster}. The triplet and triplet + OL models are our proposed triplet model as well as the outlier leveraging extension proposed. Recall is measured at $100\%$ precision.}
    \begin{tabular}{@{}llll@{}}
\toprule
{} &       Accuracy &            mAP & Recall \\
\midrule
kNN &  0.38 ± 0 & 0.10 ± 0 & 0 ± 0\\
CNN    &  0.86 ± 0.01 &  0.82 ± 0.01 &            0.03 ± 0.03 \\
Triplet      &  0.86 ± 0.00 &  0.88 ± 0.00 &            0.08 ± 0.04 \\
Triplet + OL &  0.86 ± 0.01 &  0.87 ± 0.00 &             0.05 ± 0.02 \\
\bottomrule
    \end{tabular}
    
    \label{tab:large-robustness-results}
\end{table}

We found a few things interesting here. First, a significant gap opens up between the kNN approach and the neural network approaches. The best kNN model is unable to find a good class separation. On the other hand, while the neural network-based methods required more parameters to learn, they all achieved respectable accuracy scores of around 86\%. They were also much more robust than the kNN approach. Again, we also found that the triplet loss-based approach is more robust than the CNN. However, adding outlier leveraging to the triplet loss function did not make the model more robust and even made it slightly worse. The collection strategy of $D_{large}$ was the same as $D_{tune}$, so adding data using outlier leveraging only likely helps when the extra data brings additional information on the out-of-distribution data to expect at inference. Another explanation for the gap might be that OL is primarily useful when the original dataset contains only a small number of distinct videos. Using OL in such a situation provides the model with the diversity it does not have in the original dataset.

\begin{figure}[!h]
    \centering
    \includegraphics[width=\linewidth]{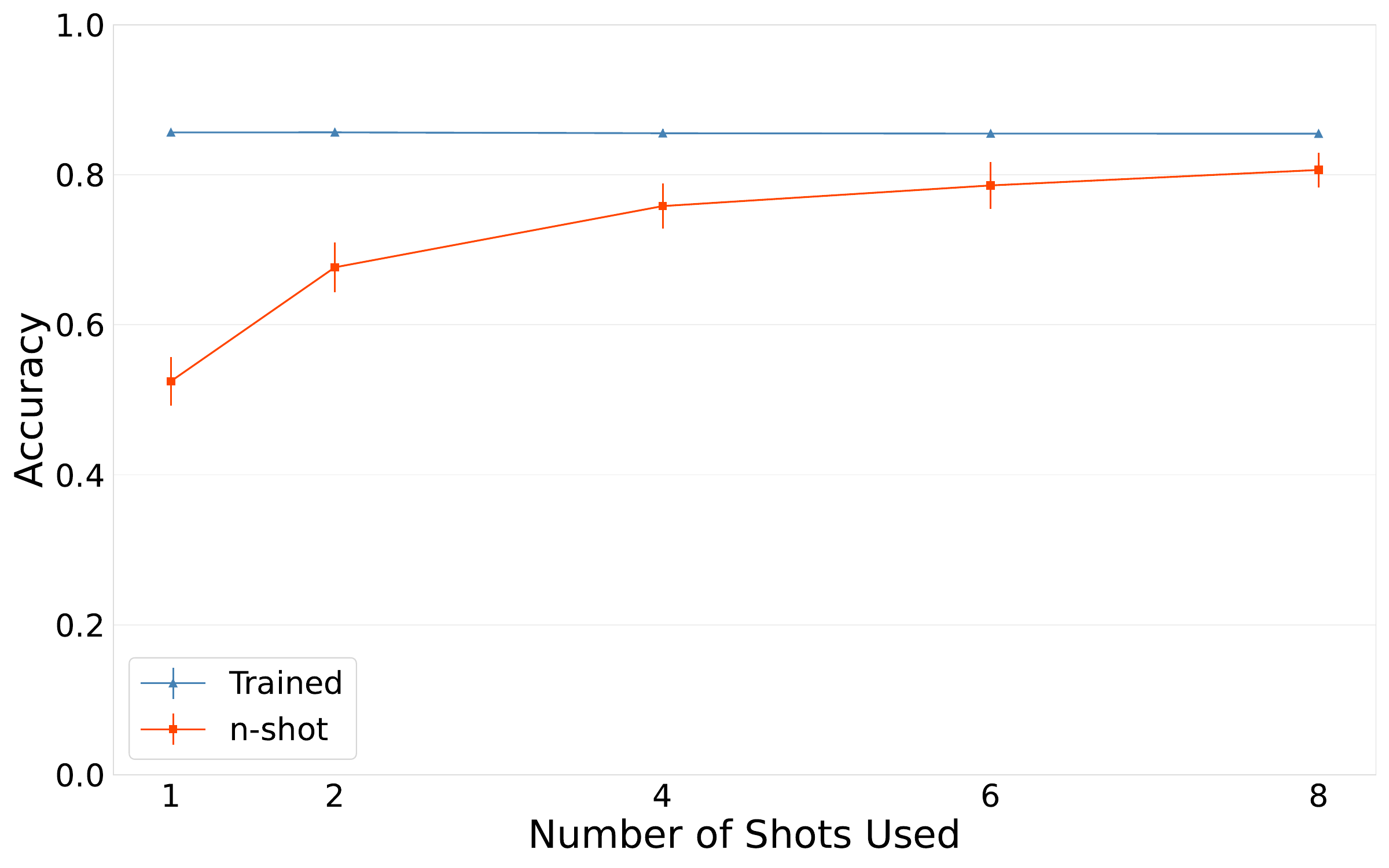}
    \caption{The accuracy for the models trained using 978 of the videos (leave out a fraction of 0.1) but where we include a variable number of shots. Using more shots increases accuracy quite drastically, achieving over $80\%$ accuracy when adding all $8$ available shots. This is despite that we do not touch the model parameters when adding these videos.}
    \label{fig:large-nshot}
\end{figure}

Given the disparity in accuracy between kNN and the triplet loss model on $D_{large}$, we performed the same experiment as in Section $6.2$. We report the results of this in Figure \ref{fig:large-nshot}. Again, we opted to leave out $10\%$ of the videos and vary the number of shots we use. Interestingly, adding more data strictly improved the accuracy of the $n$-shot videos. If we use $8$ shots to construct our centroids, we can achieve as high as $81\%$ accuracy. This is significantly higher than the kNN approach while requiring almost no compute time as we do not retrain the model. Thus, our approach can be an interesting option for quickly including new videos. 

\begin{figure}[!h]
    \centering
    \includegraphics[width=\linewidth]{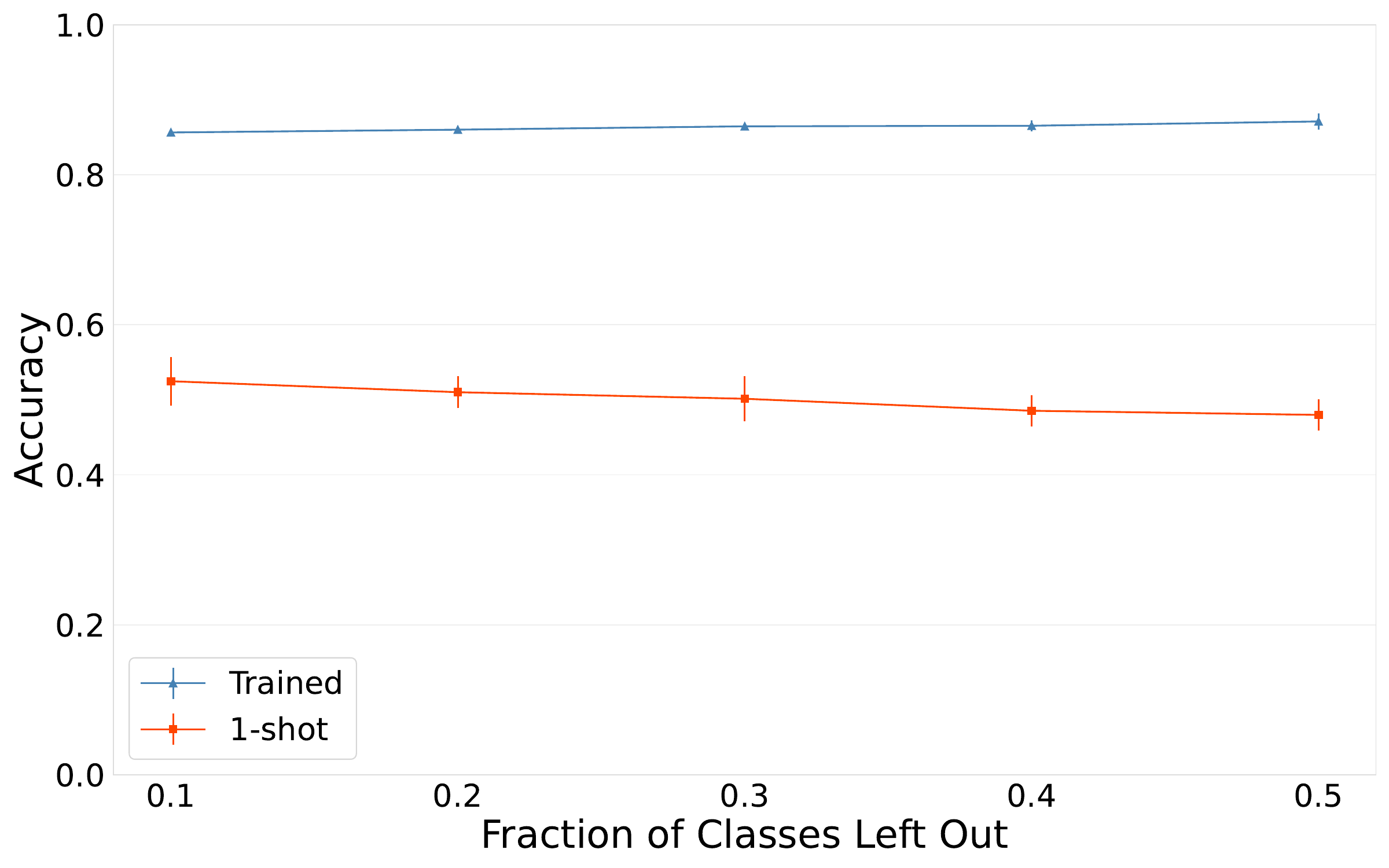}
    \caption{The accuracy for different models when varying the numbers of videos left out. A fraction of classes left out of $0.2$, for example, means that the model is trained on $870$ out of the original $1087$ videos, and the remaining $217$ are added during evaluation. Note that we do not retrain on these videos; we only add centroids for them and evaluate our models.}
    \label{fig:large-1shot}
\end{figure}
An interesting question is how far this would scale in minimal data setups. In Figure \ref{fig:large-1shot}, we vary how much data we leave out and use 1 shot for each video added at inference. What is interesting here is that the accuracy is still significantly higher than the kNN approach. When training with a leave-out fraction of $0.5$ (a $544/543$ training/inference split for video classes), the accuracy is only $2-3\%$ lower than when training with a fraction of $0.1$ (a $978/109$ training/inference split for video classes). This implies that our model can potentially incorporate arbitrarily many new videos if we have a sufficiently strong base model. This makes our approach highly scalable.


\subsection{Transferability}
As we have shown that our model can generalize to videos not seen during training, we wondered whether this extends to different settings as well. To do so, we have gathered the dataset $D_{firefox}$. This dataset consists of the same videos as $D_{large}$ but all streams were collected using the Firefox browser, which has a different implementation of the streaming controller. Thus, a model trained on Chrome is likely not transferable to Firefox. As the classification is done using centroids, we could swap the centroids of a model trained on Chrome with the centroids generated over Firefox data. 

To evaluate this hypothesis, we first took the original models we trained on $D_{large}$. For the $0$-shot setting, we use the original eight Chrome streams from $D_{large}$ to form the centroids at inference time. We dub this setting $0$-shot, as we do not use any streams from the Firefox setting. We use this setting as a baseline to evaluate how well a model trained on Chrome transfers to Firefox. For the $1$-shot and $3$-shot settings, we took our model trained on Chrome streams but evaluated it using Firefox centroids. The centroids of the $1$-shot and $3$-shot settings used one or three streams respectively. These settings demonstrate how well a model trained on Chrome streams transfers to classifying Firefox streams. Finally, we trained and evaluated our models on the Firefox data to serve as a benchmark. We report our results in Table \ref{tab:firefox-results}.

\begin{table}[t]
    \centering
        \caption{The metrics when transferring from the Chrome dataset to the Firefox dataset. The models with ($0$-shot) perform inference using the Chrome data entirely. The $1$-shot and $3$-shot settings use 1 and 3 streams from Firefox to construct the centroids. The model is trained using the Chrome streams. The models with the trained label were trained from scratch on the Firefox data. Recall is measured at $100\%$ precision.}
    \begin{tabular}{@{}llll@{}}
    \toprule
    {} &       Accuracy &            mAP & Recall \\
    \midrule
    
    CNN ($0$-shot)      &  0.07 ± 0.00 &   0.03 ± 0.01 &            0.00 ± 0.01 \\
    Triplet  ($0$-shot)       &  0.09 ± 0.01 &  0.02 ± 0.00 &                0.00 ± 0.00 \\
    Triplet ($1$-shot) &  0.25 ± 0.01 &  0.17 ± 0.01 &            0.02 ± 0.01 \\
    
    Triplet ($3$-shot) &  0.29 ± 0.01 &  0.19 ± 0.01 &            0.02 ± 0.01 \\
    kNN (Trained) &  0.12 ± 0 & 0.02 ± 0 & 0 ± 0\\
    CNN (Trained)     &  0.28 ± 0.01 &   0.13 ± 0.02 &            0.00 ± 0.01 \\
Triplet (Trained) &  0.34 ± 0.01 &  0.22 ± 0.02 &             0.03 ± 0.01 \\
    \bottomrule
    \end{tabular}

    \label{tab:firefox-results}
\end{table}

As expected, there is a transfer gap between Chrome and Firefox. Hence the setting without any shots performs poorly. What is surprising is the performance of the $1$-shot and $3$-shot transfer settings for the triplet loss model. They are both more accurate and robust than training a kNN approach entirely on Firefox data. This demonstrates that the model has learned patterns from the Chrome traffic that also generalize well to Firefox streams. Furthermore, while the accuracy of 29\% might be too low for a successful attack, it is clear from the model trained on Firefox that this setting is challenging. Nonetheless, this indicates that it is possible to train a model on streams from Chrome and use it for Firefox data as well.  

Chrome to Firefox is just one case of conversion, however. Our method is likely transferable to other settings as well. The implication of this is powerful; a sufficiently complex base model might be able to perform this side-channel attack for any arbitrary network setting. 

\section{Conclusion}

We set out to devise a side-channel attack for encrypted video streams that satisfies four properties. The attack should be robust, generalisable, scalable and transferable. Using a triplet loss-based approach alongside an extension we call outlier leveraging, we were able to address each of these issues. Our attack either beats the state-of-the-art or introduces new possibilities for the attack.

First, the triplet loss approach makes the least mistakes when dealing with out-of-distribution data. Second, it generalises well to new videos without requiring retraining. Third, these conclusions hold when the number of videos becomes large. Finally, a single model is sufficient to classify both Chrome and Firefox streams and is thus likely sufficient for any arbitrary network condition. Interesting future work would be to look further into the transferability our method permits and how to improve on it. Defences against this attack would also be a valuable angle.  

The practical implication of our work is that large-scale monitoring of video streaming content over HTTPS is likely possible. Quick solutions to alleviate the vulnerability include adjusting MPEG-DASH encoding settings to make burst patterns less identifiable, maintaining different encodings on different servers, and regularly regenerating encodings for existing videos. In the longer term, we hope that our work helps provide urgency for embedding preventative measures in the HTTPS and/or MPEG-DASH protocols to guarantee the privacy of internet users.

\bibliographystyle{IEEEtran}
\bibliography{bibliography.bib}



\end{document}